\documentclass[10pt,twocolumn,letterpaper]{article}

\usepackage{iccv}
\usepackage{times}
\usepackage{epsfig}
\usepackage{graphicx}
\usepackage{amsmath}
\usepackage{amssymb}

\usepackage{subfigure}
\usepackage{tablefootnote}
\usepackage{threeparttable}

\usepackage{pdfpages}

\usepackage[pagebackref=true,breaklinks=true,letterpaper=true,colorlinks,bookmarks=false]{hyperref}

\iccvfinalcopy 

\def\iccvPaperID{4449} 

\ificcvfinal\fi
\begin{document}

\title{Recursive Cascaded Networks for Unsupervised Medical Image Registration}

\author{Shengyu Zhao\textsuperscript{2,3} \quad Yue Dong\textsuperscript{2} \quad Eric I-Chao Chang\textsuperscript{3} \quad Yan Xu\textsuperscript{1,3}\thanks{Corresponding author. This work is supported by the National Science and Technology Major Project of the Ministry of Science and Technology in China under Grant 2017YFC0110903,  Microsoft Research under the eHealth program, the National Natural Science Foundation in China under Grant 81771910, the Fundamental Research Funds for the Central Universities of China under Grant SKLSDE-2017ZX-08 from the State Key Laboratory of Software Development Environment in Beihang University in China, the 111 Project in China under Grant B13003.} \\\\
\textsuperscript{1}School of Biological Science and Medical Engineering and Beijing Advanced \\
Innovation Centre for Biomedical Engineering, Beihang University \\
\textsuperscript{2}IIIS, Tsinghua University \\
\textsuperscript{3}Microsoft Research \\
{\tt\small zsyzzsoft@gmail.com, dongyue8@gmail.com, echang@microsoft.com, xuyan04@gmail.com}
}


\maketitle
\ificcvfinal\thispagestyle{empty}\fi

\begin{abstract}
We present recursive cascaded networks, a general architecture that enables learning deep cascades, for deformable image registration.
The proposed architecture is simple in design and can be built on any base network.
The moving image is warped successively by each cascade and finally aligned to the fixed image; this procedure is recursive in a way that every cascade learns to perform a progressive deformation for the current warped image. The entire system is end-to-end and jointly trained in an unsupervised manner. In addition, enabled by the recursive architecture, one cascade can be iteratively applied for multiple times during testing, which approaches a better fit between each of the image pairs. We evaluate our method on 3D medical images, where deformable registration is most commonly applied. We demonstrate that recursive cascaded networks achieve consistent, significant gains and outperform state-of-the-art methods. The performance reveals an increasing trend as long as more cascades are trained, while the limit is not observed. Code is available at \url{https://github.com/microsoft/Recursive-Cascaded-Networks}.

\end{abstract}

\section{Introduction}

\begin{figure*}[t]
\begin{center}
   \includegraphics[width=0.95\linewidth]{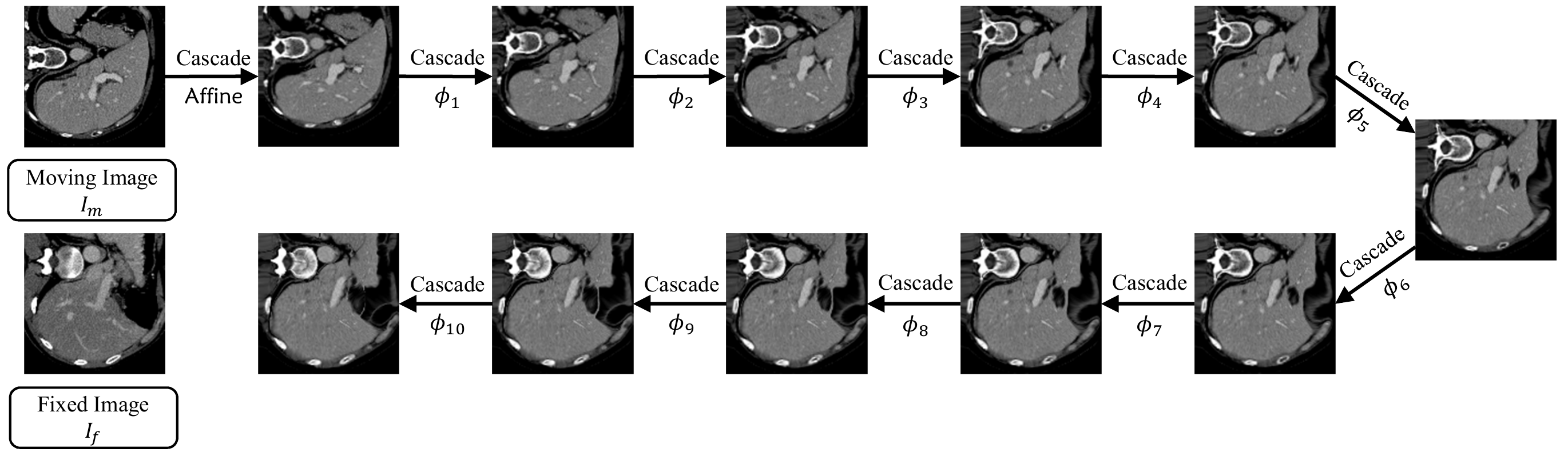}
\end{center}
  \caption{Example of recursive cascaded networks for registering liver CT scans. The moving image is recursively and progressively warped by each of the cascades, finally aligned to the fixed image. Each $\phi_k$ denotes a predicted flow field, taken both the preceding warped image and the fixed image as inputs. Only image slices are presented but note that the registration is actually performed in 3D.}
\label{fig:example}
\end{figure*}

\begin{figure}[t]
\begin{center}
   \includegraphics[width=1.0\linewidth]{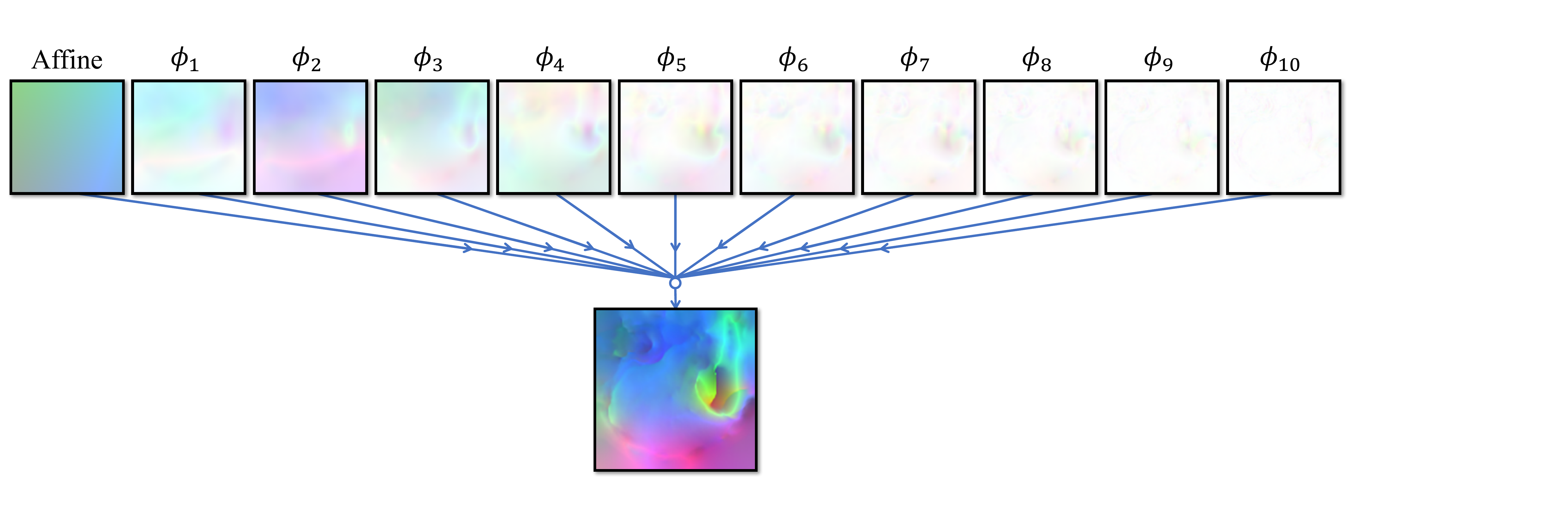}
\end{center}
  \caption{Composition of flow fields, corresponding to the example shown in Figure \ref{fig:example}. The final flow prediction is composed of an initial affine transformation and $\phi_1, \dots, \phi_n$, each of which only performs a rather simple displacement. We can see that the top cascades mainly learn a global alignment, while the bottom cascades play a role of refinement. Flow fields are drawn by mapping the abosolute value of the three components ($x, y, z$) of flow displacements into color channels (R, G, B) respectively. White area indicates zero displacement.}
\label{fig:cascade_flow}
\end{figure}

Deformable image registration has been studied in plenty of works and raised great importance. The non-linear correspondence between a pair of images is established by predicting a deformation field under the smoothness constraint. Among traditional algorithms, an iterative approach is commonly suggested~\cite{TraFast,TraMethods,SyN,TraElastic,TraLarge,TraSTBrain,TraLP,TraHAMMER}, where they formulate each iteration as a progressive optimization problem.

Image registration has drawn growing interests in terms of deep learning techniques. A closely related area is optical flow estimation, which is essentially a 2D image registration problem but the flow fields are discontinuous across objects and the tracking is mainly about motion with rare color difference. Occlusions and folding areas requiring a guess are inevitable in optical flow estimation (but not expected in deformable image registration). Automatically generated datasets (e.g., Flying Chairs~\cite{Flownet}, Flying Things 3D~\cite{FlyingThings3D}) are of great help for supervising convolutional neural networks (CNNs) in such settings~\cite{Flownet,LiteFlownet,Flownet2.0,PwcNet,DCFlow}. Some studies also try to stack multiple networks. They assign different tasks and inputs to each cascade in a non-recursive way and train them one by one~\cite{Flownet2.0,ResFlownet}, but their performance approaches a limit with only a few (no more than 3) cascades. On the other hand, cascading may not help much when dealing with discontinuity and occlusions. Thus by intuition, we suggest that cascaded networks with a recursive architecture fits the setting of deformable registration.

Learning-based methods are also suggested as an approach in deformable image registration. Unlike optical flow estimation, intersubject registration with vague correspondence of image intensity is usually demanded. Some initial works rely on the dense ground-truth flows obtained by either traditional algorithms~\cite{SupCNNSyN,SupRegFast} or simulating intrasubject deformations~\cite{SupRegIntra,SupRegCNN}, but their performance is restricted due to the limited quality of training data.

Unsupervised learning methods with comparable performance to traditional algorithms have been presented recently~\cite{VM,ExVM,DLIR,SimpleDLIR,VTN,RegFCN}. They only require a similarity measurement between the warped moving image and the fixed image, while the gradients can backpropagate through the differentiable warping operation (a.k.a.\@ spatial transformer~\cite{STN}). However, most proposed networks are enforced to make a straightforward prediction, which proves to be a burden when handling complicated deformations especially with large displacements. DLIR~\cite{DLIR} and VTN~\cite{VTN} also stack their networks, though both limited to a small number of cascades. DLIR trains each cascade one by one, i.e., after fixing the weights of previous cascades. VTN jointly trains the cascades, while all successively warped images are measured by the similarity compared to the fixed image. Neither training method allows intermediate cascades to progressively register a pair of images. Those non-cooperative cascades learn their own objectives regardless of the existence of others, and thus further improvement can hardly be achieved even if more cascades are conducted. They may realize that network cascading possibly solves this problem, but there is no effective way of training deep network cascades for progressive alignments.

Therefore, we propose the recursive cascade architecture, which encourages the unsupervised training of an unlimited number of cascades that can be built on existing base networks, for advancing the state of the art. The difference between our architecture and existing cascading methods is that each of our cascades commonly takes the current warped image and the fixed image as inputs (in contrast to~\cite{Flownet2.0,ResFlownet}) and the similarity is only measured on the final warped image (in contrast to~\cite{DLIR,VTN}), enabling all cascades to learn progressive alignments cooperatively. Figure \ref{fig:example} shows an example of applying the proposed architecture built on 10 deformable cascades of the base network VTN.

Conceptually, we formulate the registration problem as determining a parameterized flow prediction function, which outputs a dense flow field based on the input of an image pair. This function can be recursively defined on the warped moving image with essentially the same functionality. Instead of training the function in a straightforward way, the final prediction can be considered a composition of recursively predicted flow fields, while each cascade only needs to learn a simple alignment of small displacement that can be refined by deeper recursion. Figure \ref{fig:cascade_flow} verifies our conception. Our method also enables the use of shared-weight cascades, which potentially achieves performance gains without introducing more parameters.

To summarize, we present a deep recursive cascade architecture for deformable image registration, which facilitates the unsupervised end-to-end learning and achieves consistent gains independently of the base network; shared-weight cascading technique with direct test-time improvement is developed as well. We conduct extensive experiments based on diverse evaluation metrics (segmentations and landmarks) and multiple datasets across image types (liver CT scans and brain MRIs).




\begin{figure*}[t]
\begin{center}
   \includegraphics[width=0.9\linewidth]{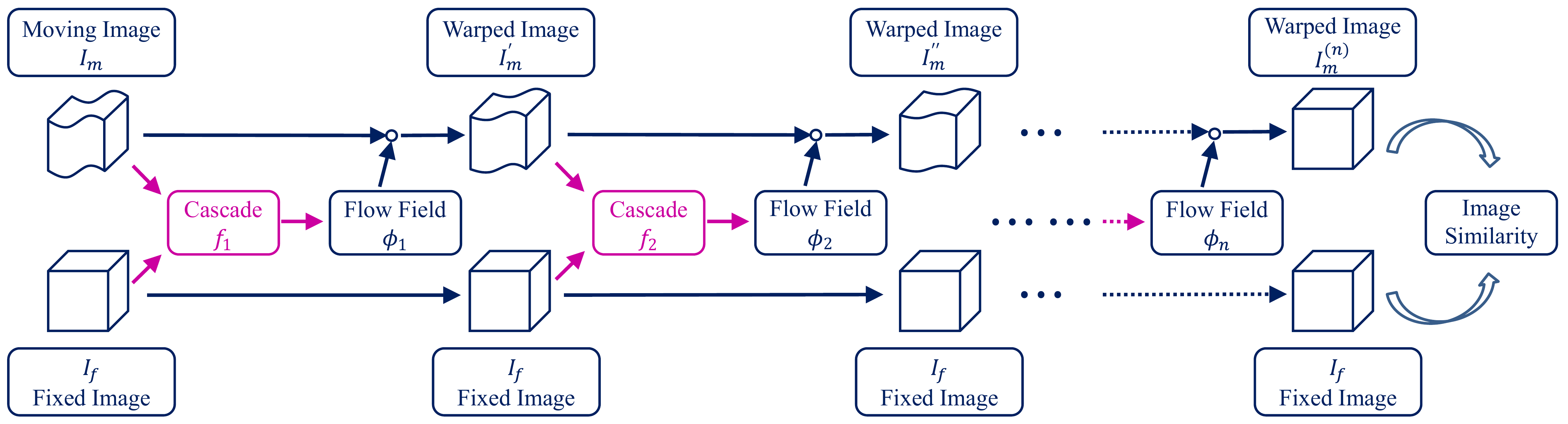}
\end{center}
   \caption{Illustration of our recursive cascade architecture. Circle denotes a composition, where the preceding warped image ($I_m^{(k-1)}$) is reconstructed by the predicted flow field ($\phi_k$), resulting in the successive warped image ($I_m^{(k)}$). The unsupervised end-to-end learning is only guided by the image similarity between $I_m^{(n)}$ and $I_f$, in contrast to previous works.}
\label{fig:flowchart}
\end{figure*}

\section{Related Work}

Cascade approaches have been involved in a variety of domains of computer vision, e.g., cascaded pose regression progressively refines a pose estimation learned from supervised training data~\cite{CPR}, cascaded classifiers speed up the process of object detection~\cite{CascadeDetection}.

Deep learning also benefits from cascade architectures. For example, deep deformation network~\cite{DeepDeformLandmark} cascades two stages and predicts a deformation for landmark localization. Other applications include object detection~\cite{CascadeRCNN}, semantic segmentation~\cite{CascadeSeg}, and image super-resolution~\cite{CascadeSR}. There are also several works specified to medical images, e.g., 3D image reconstruction for MRIs~\cite{Cascade7TMRI,CascadeMR}, liver segmentation~\cite{ShapeSeg} and mitosis detection~\cite{CascadeMedicalDetection}. Note that shallow, non-recursive network cascades are usually proposed in those works.

In respect of registration, traditional algorithms iteratively optimize some energy functions in common~\cite{TraFast,TraMethods,SyN,TraElastic,TraLarge,TraSTBrain,TraLP,TraHAMMER}. Those methods are also recursive in general, i.e., similarly functioned alignments with respect to the current warped images are performed during iterations. Iterative Closest Point is an iterative, recursive approach for registering point clouds~\cite{ICP1,ICP2}, where the closest pairs of points are matched at each iteration and a rigid transformation that minimizes the difference is solved. In deformable image registration, most traditional algorithms basically works like this but in a much more complex way. Standard symmetric normalization (SyN)~\cite{SyN} maximizes the cross-correlation within the space of diffeomorphic maps during iterations. Optimizing free-form deformations using B-spline~\cite{BSpline} is another standard approach.

Learning-based methods are presented recently. Supervised methods entail much effort on the labeled data that can hardly meet the realistic demands, resulting in the limited performance~\cite{SupCNNSyN,SupRegFast,SupRegIntra,SupRegCNN}. Unsupervised methods are proposed to solve this problem. Several initial works shows the possibility of unsupervised learning~\cite{DLIR,SimpleDLIR,RegFCN,DL4DCT}, among which DLIR~\cite{SimpleDLIR} performs on par with the B-spline method implemented in SimpleElastix~\cite{SimpleElastix} (a multi-language extension of Elastix~\cite{Elastix}, which is selected as one of our baseline methods). VoxelMorph~\cite{VM} and VTN~\cite{VTN} achieve better performance by predicting a dense flow field using deconvolutional layers~\cite{Deconv}, whereas DLIR only predicts a sparse displacement grid interpolated by a third order B-spline kernel. VoxelMorph only evaluates their method on brain MRI datasets~\cite{VM,ExVM}, but shown deficiency on other datasets such as liver CT scans by later work~\cite{VTN}. Additionally, VTN proposes an initial convolutional network which performs an affine transformation before predicting deformation fields, leading to a truly end-to-end framework by substituting the traditional affine stage.

State-of-the-art VTN and VoxelMorph are selected as our base networks, and the suggested affine network is also integrated as our top-level cascade. To our knowledge, none of those work realizes that training deeper cascades advances the performance for deformable image registration.

\section{Recursive Cascaded Networks}

Let $I_m, I_f$ denote the moving image and the fixed image respectively, both defined over $d$-dimensional space $\Omega$. A flow field is a mapping $\phi: \Omega \to \Omega$. For deformable image registration, a reasonable flow field should be continuously varying and prevented from folding. The task is to construct a flow prediction function $F$ which takes $I_m, I_f$ as inputs and predicts a dense flow field that aligns $I_m$ to $I_f$.

We cascade this procedure by recursively performing registration on the warped image. The warped image $I_m'$ is exactly the composition of the flow field and the moving image, namely
\begin{align}
    I_m' = \phi \circ I_m.
\end{align}
Conceptually,
\begin{align}
    F(I_m, I_f) = \phi \circ F_1(\phi \circ I_m, I_f),
\end{align}
where $F_1$ may be the same as $F$, but in general a different flow prediction function. This recursion can be infinitely applied in theory.

Following this recursion, the moving image is warped successively, enabling the final prediction (probably with large displacement) to be decomposed into cascaded, progressive refinements (with small displacements). One cascade is basically a flow prediction function ($f_k$), and the $k$-th cascade predicts a flow field of
\begin{align}
    \phi_k = f_k(I_m^{(k-1)}, I_f).
\end{align}
$I_m^{(k)}$ denotes the moving image warped by the first $k$ cascades. Figure \ref{fig:flowchart} depicts the proposed architecture. Assuming for $n$ cascades in total, the final output is a composition of all predicted flow fields, i.e.,
\begin{align}
    F(I_m, I_f) = \phi_{n} \circ \cdots \circ \phi_1,
\end{align}
and the final warped image is constructed by
\begin{align}
    I_m^{(n)} = F(I_m, I_f) \circ I_m.
\end{align}

\subsection{Subnetworks}
Each $f_k$ is implemented as a convolutional neural network in this paper. Every network is designed to predict a deformable flow field on itself based on the input warped image and the fixed image. $f_1, \dots f_n$ can be different in network architecture, but surely using a common base network is well-designed enough for convenience. Those cascades may learn different network parameters on each, since one cascade is allowed to learn a part of measurements or perform some type of alignment specifically. Note that the images input to the networks are discretized and so are the output flow fields, thus we treat them by multilinear interpolation (or simply trilinear interpolation for 3D images), and out-of-bound indices by nearest-point interpolation~\cite{VTN}.

An architecture similar to the U-Net~\cite{ImageTransUNet,UNet} is widely used for deformable registration networks, such as VTN~\cite{VTN} and VoxelMorph~\cite{VM}. Such network consists of encoders followed by decoders with skip connections. The encoders help to extract features, while the decoders perform upsampling and refinement, ending with a dense prediction.

For medical images, it is usually the case that two scans can be roughly aligned by an initial rigid (or affine) transformation. VoxelMorph~\cite{VM} assumes that input images are pre-affined by an external tool, whereas VTN~\cite{VTN} integrates an efficient affine registration network which outperforms the traditional stage. As a result, we also embed the affine registration network as our top-level cascade, which behaves just like a normal one except that it is only allowed to predict an affine transformation rather than general flow fields.

\subsection{Unsupervised End-to-End Learning}
We suggest that all cascades can be jointly trained by merely measuring the similarity between $I_m^{(n)}$ and $I_f$ together with regularization losses. Enabled by the differentiable composition operator (i.e., warping operation), recursive cascaded networks can learn to perform progressive alignments cooperatively without supervision. To our knowledge, no previous work achieves good performance by stacking more than 3 deformable registration networks, partly because they train them one by one~\cite{DLIR} (then the performance can hardly improve) or they measure the similarity on each of the warped images~\cite{VTN} (then the networks can hardly learn progressive alignments).

Regularization losses are basically the smooth terms of $\phi_1, \dots, \phi_n$, and thus are necessary. Every predicted flow field is penalized by an L2 variation loss as done in~\cite{VM,VTN}. The affine cascade works with its own regularization losses introduced in VTN~\cite{VTN}.

\subsection{Shared-Weight Cascading}

One cascade can be repetitively applied during recursion. I.e., multiple cascades can be shared with the same parameters, and that is called shared-weight cascading.

After an $n$-cascade network is trained, we can still possibly apply additional shared-weight cascades during testing. For example, we may replicate all cascades as an indivisible whole by the end of $I_m^{(n)}$, i.e., totally $2n$ cascades are associated with flow prediction functions $f_1, \dots, f_n, f_1, \dots, f_n$ respectively. We develop a better approach by immediately inserting one or more shared-weight cascades after each, i.e., totally $r$$\times$$n$ cascades are constructed by substituting each $f_k$ by $r$ times of that. This approach will be proved to be effective later in the experiments.

Shared-weight cascading during testing is an option when the quality of output flow fields can be improved by further refinement. However, we note that this technique does not always get positive gains and may lead to over deformation. Recursive cascades only ensure an increasing similarity between the warped moving image and the fixed image, but the aggregate flow field becomes less natural if the images are too perfectly matched.


The reason we do not use shared-weight cascading in training is that shared-weight cascades consume extra GPU memory as large as non-shared-weight cascades during gradient backpropagation in the platform we use (Tensorflow~\cite{Tensorflow}). The number of cascades to train is constrained by the GPU memory, but they would perform better with the allowance of learning different parameters when the dataset is large enough to avoid overfitting. 

\section{Experiments}

\subsection{Experimental Settings}

We build our recursive cascaded networks mainly based on the network architecture of VTN~\cite{VTN}, which is a state-of-the-art method for deformable image registration. Note that VTN already stacks a few cascades of their deformable subnetworks, and a single cascade is being used as our base network. Up to 10-cascade VTN (excluding the affine cascade) is jointly trained using our proposed method. To show the generalizability of our architecture, we also choose VoxelMorph~\cite{ExVM} as another base network. We train up to 5-cascade VoxelMorph, because each cascade of VoxelMorph consumes more resources.

We evaluate our method on two types of 3D medical images: liver CT scans and brain MRI scans. For liver CT scans, we train and test recursive cascaded networks for pairwise, subject-to-subject registration, which stands for a general purpose of allowing the fixed image to be arbitrary. For brain MRI scans, we follow the experimental setup of VoxelMorph~\cite{VM}, where each moving image is registered to a fixed atlas, called atlas-based registration. Both settings are common in medical image registration.

\paragraph{Implementation.}

Inherited from the implementation of VTN~\cite{VTN} using Tensorflow 1.4~\cite{Tensorflow} built with a custom warping operation, the correlation coefficient is used as the similarity measurement, while the ratios of regularization losses are kept the same as theirs. We train our model using a batch size of $4$, on 4 cards of 12G NVIDIA TITAN Xp GPU. The training stage runs for $10^5$ iterations with the Adam optimizer~\cite{Adam}. The learning rate is initially $10^{-4}$ and halved after $6 \times 10^4$ steps and again after $8 \times 10^4$ steps.

\paragraph{Baseline Methods.}

VTN~\cite{VTN} and VoxelMorph~\cite{VM} are state-of-the-art learning-based methods. We cascade their base networks and also compare with the original systems. Besides, we also compare against SyN~\cite{SyN} (integrated in ANTs~\cite{ANTs} together with the affine stage) and B-spline~\cite{BSpline} (integrated in Elastix~\cite{Elastix} together with the affine stage), which are shown to be the top-performing traditional methods for deformable image registration~\cite{VM,EvalDeform,VTN}. We run ANTs SyN and Elastix B-spline with the parameters recommended in VTN~\cite{VTN}.

\paragraph{Evaluation Metrics.}

We quantify the performance by the Dice score~\cite{Dice} based on the segmentation of some anatomical structure, between the warped moving image and the fixed image, as done in~\cite{VM,DLIR}. The Dice score of two regions $A, B$ is formulated as
\begin{align}
    \mathrm{Dice}(A, B) = 2 \cdot \frac{|A \cap B|}{|A| + |B|}.
\end{align}
Perfectly overlapped regions come with a Dice score of $1$. The Dice score explicitly measures the coincidence between two regions and thereby reflects the quality of registration. If multiple anatomical structures are annotated, we compute the Dice score with respect to each and take an average.

In addition, landmark annotations are available in some datasets and can be utilized as an auxiliary metric. We compute the average distance between the landmarks of the fixed image and the warped landmarks of the moving image, also introduced in VTN~\cite{VTN}.


\begin{table*}[t]
\small
\centering
\begin{threeparttable}
\begin{tabular}{|c||cc|c|c||c||cc|}
\hline
Method                          & \multicolumn{2}{c|}{SLIVER}  & LiTS & LSPIG& LPBA      & \multicolumn{2}{c|}{Time (sec)} \\
                                & Dice          & Lm.\@ Dist.     & Dice          & Dice          & Avg.\@ Dice          & GPU   & CPU   \\
\hline\hline
ANTs SyN~\cite{SyN,ANTs}        & 0.895 (0.037) & 12.2 (5.7)    & 0.862 (0.055) & 0.825 (0.063) & 0.708 (0.015) & -     & 748   \\
Elastix B-spline~\cite{Elastix,BSpline}          & 0.910 (0.038) & 12.6 (6.6)    & 0.863 (0.059) & 0.825 (0.059) & 0.675 (0.013) & -     & 115 \\
\hline
VoxelMorph\tnote{1} ~\cite{ExVM} & 0.883 (0.034) & 14.0 (4.6) & 0.831 (0.061) & 0.715 (0.090) & 0.685 (0.017) & 0.20 & 17\\
VoxelMorph (reimplem.)\tnote{2}  & 0.913 (0.025) & 13.1 (4.7) & 0.870 (0.048) & 0.833 (0.057) & 0.688 (0.015) & 0.15 & 14 \\
5-cascade VoxelMorph            & 0.944 (0.017) & 12.4 (4.9) & 0.903 (0.055) & 0.849 (0.062) & 0.708 (0.015) & 0.41 & 69 \\
3$\times$5-cascade VoxelMorph   & 0.950 (0.014) & 11.9 (4.9) & 0.905 (0.065) & 0.842 (0.066) & 0.715 (0.014) & 1.09 & 201\\
\hline
VTN (ADDD)\tnote{3} ~\cite{VTN}           & 0.942 (0.020) & 12.0 (4.9) & 0.897 (0.049) & 0.846 (0.064) & 0.701 (0.014) & 0.13 & 26\\
10-cascade VTN                  & 0.953 (0.014) & 10.8 (4.9) & \textbf{0.909 (0.060)} & \textbf{0.855 (0.060)} & 0.716 (0.013) & 0.25 & 87 \\
2$\times$10-cascade VTN         & \textbf{0.956 (0.012)} & \textbf{10.2 (4.7)} & 0.908 (0.070) & 0.849 (0.063) & \textbf{0.719 (0.012)} & 0.42 & 179 \\
\hline
\end{tabular}
\caption{Comparison among traditional methods (ANTs SyN and Elastix B-spline), our baseline networks (VoxelMorph and VTN), and our proposed recursive cascaded networks with and without shared-weight cascading. $r$$\times$$n$-cascade means that every deformable cascade is repetitively applied for $r$ times during testing, using our proposed shared-weight cascading method. For liver datasets (SLIVER, LiTS, and LSPIG), the Dice score measures the overlap of liver segmentations, and Lm.\@ Dist.\@ means an average distance among 6 annotated landmarks. Avg.\@ Dice means an average Dice score among all 56 segmented anatomical structures for the brain dataset LPBA. Standard deviations across instances are in parentheses.}
\label{table:method}
\begin{tablenotes}
\item[1] Images for training and testing are pre-affined (as required in VoxelMorph~\cite{ExVM}) using ANTs~\cite{ANTs}.
\item[2] Reimplemented with an integrated affine network and trained using our method.
\item[3] Denotes one affine registration subnetwork plus three dense deformable subnetworks~\cite{VTN}.
\end{tablenotes}
\end{threeparttable}
\end{table*}

\subsection{Datasets}

For liver CT scans, we use the following datasets:
\begin{itemize}
\item MSD~\cite{MSD}. This dataset contains various types of medical images for segmenting different target objects. CT scans of liver tumours (70 scans excluding LiTS), hepatic vessels (443 scans), and pancreas tumours (420 scans) are selected since liver is likely to be included.

\item BFH (introduced in VTN~\cite{VTN}), 92 scans.

\item SLIVER~\cite{MICCAI07}, 20 scans with liver segmentation ground truth. Additionally, 6 anatomical keypoints selected as landmarks are annotated by 3 expert doctors, and we take their average as ground truth.

\item LiTS~\cite{LiTS}, 131 scans with liver segmentation ground truth.

\item LSPIG (Liver Segmentation of Pigs, provided by the First Affiliated Hospital of Harbin Medical University), 
containing 17 pairs of CT scans from pigs, along with liver segmentation ground truth. Each pair comes from one pig with (perioperative) and without (preoperative) 13 mm Hg pneumoperitoneum pressure.

\end{itemize}

Unsupervised methods are trained on the combination of MSD and BFH with $1025^2$ ($1025 = 70 + 443 + 420 + 92$) image pairs in total. SLIVER ($20 \times 19$ image pairs) and LiTS ($131 \times 130$ image pairs) are used for regular evaluation, while LSPIG is regarded as a challenging dataset which entails generalizability. Only 34 intrasubject image pairs in LSPIG, each of which comes from a same pig (preoperative to perioperative, or vice versa), are evaluated.

For brain MRI scans, we use the following datasets:
\begin{itemize}
\item ADNI
~\cite{ADNI}, 66 scans.

\item ABIDE
~\cite{ABIDE}, 1287 scans.

\item ADHD
~\cite{ADHD}, 949 scans.

\item LPBA (LONI Probabilistic Brain Atlas)~\cite{LPBA40}. This dataset contains 40 scans, each of which comes with segmentation ground truth of 56 anatomical structures.

\end{itemize}

ADNI, ABIDE, ADHD are used for training, and LPBA for testing. All 56 anatomical structures are evaluated by an average Dice score. For atlas-based registration, the first scan in LPBA is fixed as the atlas in our experiments, which is shown to be without loss of generality later in the atlas analysis.

We carry out standard preprocessing steps referring to VTN~\cite{VTN} and VoxelMorph~\cite{VM}. Raw scans are resampled into $128 \times 128 \times 128$ voxels after cropping unnecessary area around the target object. For liver CT scans, a simple threshold-based algorithm is applied to find a rough liver bounding box for cropping. For brain MRI scans, skulls are first removed using FreeSurfer~\cite{FreeSurfer}. The volumes are visualized for quality control so that seldom badly processed images are manually removed. (An overview of the evaluation datasets is provided in the supplementary material.)

\subsection{Results}

Table \ref{table:method} summarizes our overall performance compared with state-of-the-art methods. Running times are approximately the same across datasets, so we test them on SLIVER, with an NVIDIA TITAN Xp GPU and an Intel Xeon E5-2690 v4 CPU. No GPU implementation of ANTs or Elastix has been found, nor in previous works~\cite{ANTs,VM,DLIR,Elastix,VTN}. Figure \ref{fig:experiment} visualizes those methods on an example in the brain dataset LPBA. (See the supplementary material for more examples.)

As shown in Table \ref{table:method}, recursive cascaded networks outperform the existing methods in all our datasets with significant gains. More importantly, the proposed architecture is independent of the base network, not limited to VTN~\cite{VTN} and VoxelMorph~\cite{VM}. Although the number of cascades causes linear increments to the running times, a 10-cascade VTN still runs in a comparable (GPU) time to the baseline networks, showing the efficiency of our architecture.

\begin{figure}[t]
\begin{center}
   \includegraphics[width=1.0\linewidth]{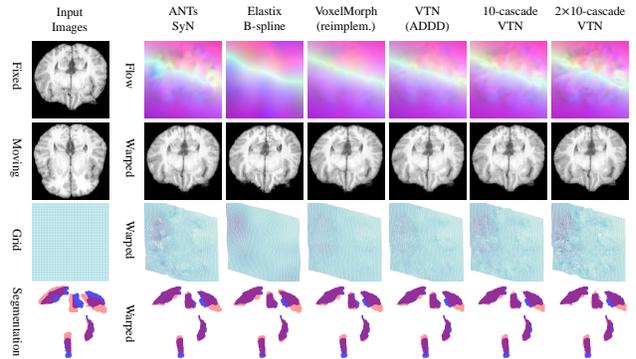}
\end{center}
   \caption{Visualization of an example in the brain dataset LPBA. Grids of deeper color represent lower height. Segmentations of 5 chosen anatomical structures are presented by projection. Blue areas stand for the segmentations of the fixed image, and red areas for the moving image or the warped images.}
\label{fig:experiment}
\end{figure}

\begin{table*}[t]
\small
\centering
\begin{threeparttable}
\begin{tabular}{|c||cc|c|c||c||cc|}
\hline
Architecture & \multicolumn{2}{c|}{SLIVER}  & LiTS          & LSPIG & LPBA & \multicolumn{2}{c|}{Time (sec)}\\
      & Dice          & Lm.\@ Dist.     & Dice          & Dice  & Avg.\@ Dice & GPU & CPU \\
\hline\hline
Affine only & 0.794 (0.042) & 14.8 (4.7) & 0.754 (0.059) & 0.727 (0.054) & 0.628 (0.017) & 0.08 & 0.4 \\
\hline
1-cascade VoxelMorph & 0.913 (0.025) & 13.1 (4.7) & 0.867 (0.050) & 0.833 (0.057) & 0.688 (0.015) & 0.15 & 14 \\
2-cascade VoxelMorph & 0.933 (0.021) & 12.8 (4.8) & 0.888 (0.048) & 0.845 (0.057) & 0.699 (0.014) & 0.21 & 27 \\
3-cascade VoxelMorph & 0.940 (0.018) & 12.6 (5.0) & 0.897 (0.049) & 0.849 (0.060)  & 0.706 (0.014) & 0.28 & 40 \\
4-cascade VoxelMorph & 0.943 (0.017) & 12.5 (5.1) & 0.900 (0.052) & \textbf{0.851 (0.058)} & 0.707 (0.014) & 0.35 & 54 \\
5-cascade VoxelMorph & \textbf{0.944 (0.017)} & \textbf{12.4 (4.9)} & \textbf{0.903 (0.055)} & 0.849 (0.062) & \textbf{0.708 (0.015)} & 0.41 & 69 \\
\hline
1-cascade VTN & 0.914 (0.025) & 13.0 (4.8) & 0.870 (0.048) & 0.833 (0.054) & 0.686 (0.014) & 0.10 & 10 \\
2-cascade VTN & 0.935 (0.020) & 12.2 (4.7) & 0.891 (0.045) & 0.843 (0.061) & 0.697 (0.014) & 0.12 & 18 \\
3-cascade VTN & 0.943 (0.018) & 11.8 (4.7) & 0.900 (0.045) & 0.850 (0.060) & 0.703 (0.014) & 0.13 & 26 \\
4-cascade VTN & 0.948 (0.016) & 11.6 (4.8) & 0.906 (0.047) & 0.852 (0.063) & 0.708 (0.014) & 0.15 & 35 \\
5-cascade VTN & 0.949 (0.015) & 11.5 (4.8) & 0.908 (0.051) & 0.853 (0.064) & 0.709 (0.014) & 0.17 & 47 \\
6-cascade VTN & 0.951 (0.015) & 11.3 (4.9) & \textbf{0.910 (0.050)} & 0.852 (0.064) & 0.712 (0.014) & 0.18 & 57 \\
7-cascade VTN & 0.951 (0.015) & 11.2 (4.9) & 0.908 (0.055) & 0.852 (0.061) & 0.712 (0.013) & 0.20 & 65\\
8-cascade VTN & 0.952 (0.014) & 11.1 (4.7) & \textbf{0.910 (0.056)} & 0.854 (0.059) & 0.714 (0.013) & 0.22 & 75\\
9-cascade VTN & \textbf{0.953 (0.014)} & 10.9 (4.7) & \textbf{0.910 (0.059)} & 0.851 (0.064) & \textbf{0.716 (0.013)} & 0.23 & 90 \\
10-cascade VTN & \textbf{0.953 (0.014)} & \textbf{10.8 (4.9)} & 0.909 (0.060) & \textbf{0.855 (0.060)} & \textbf{0.716 (0.013)} & 0.25 & 87 \\
\hline
\end{tabular}
\caption{Comparison among different number of recursive cascades. $n$-cascade means $n$ recursive cascades of the base network, excluding the affine cascade. Standard deviations across instances are in parentheses.}
\label{table:cascade}
\end{threeparttable}
\end{table*}

\begin{figure*}[t]
\begin{center}
\subfigure[Dice scores on liver datasets.]{ \label{liver_dice}
   \includegraphics[width=0.32\linewidth]{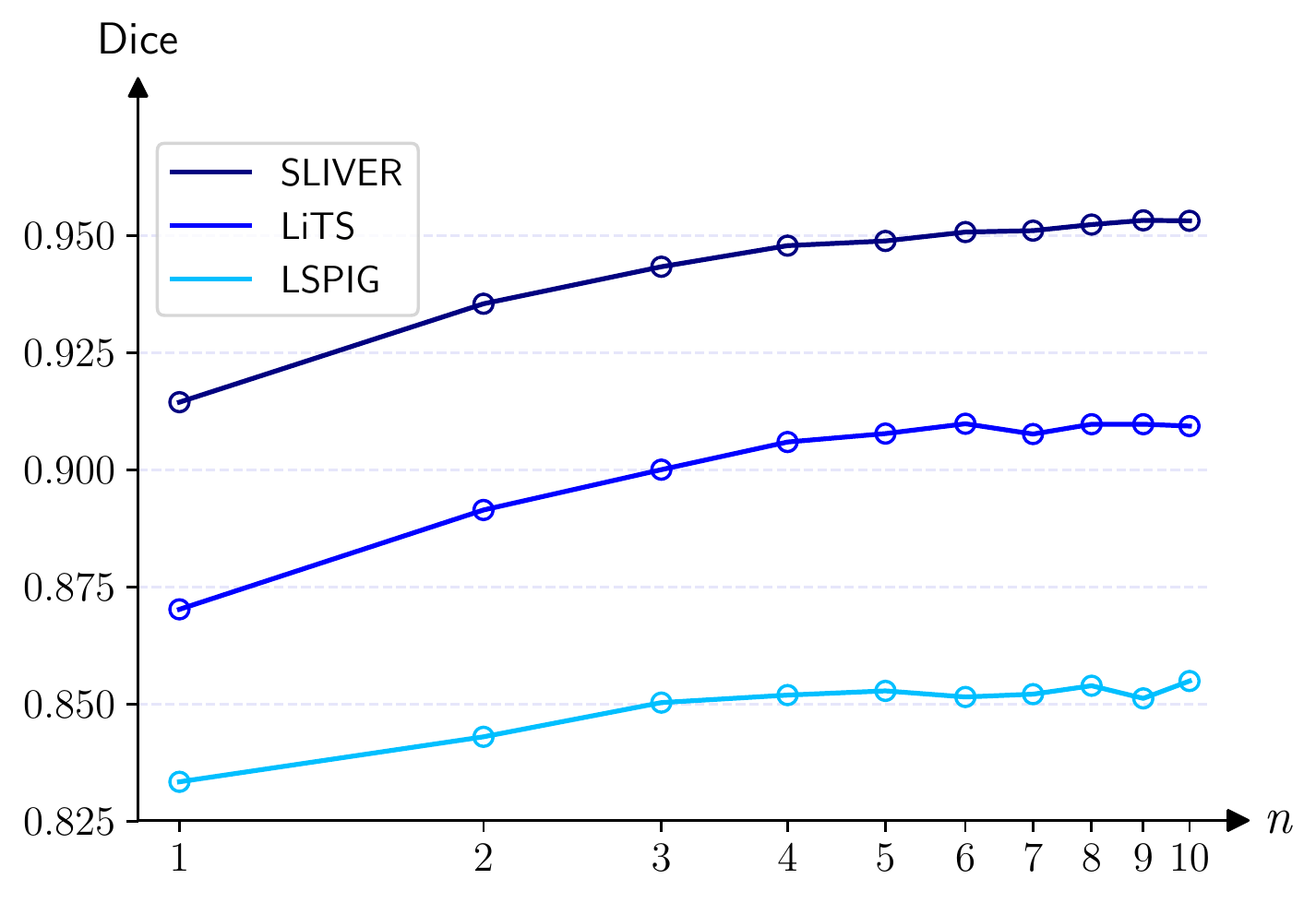}}
\subfigure[Landmark distances on SLIVER.]{ \label{liver_dist}
   \includegraphics[width=0.32\linewidth]{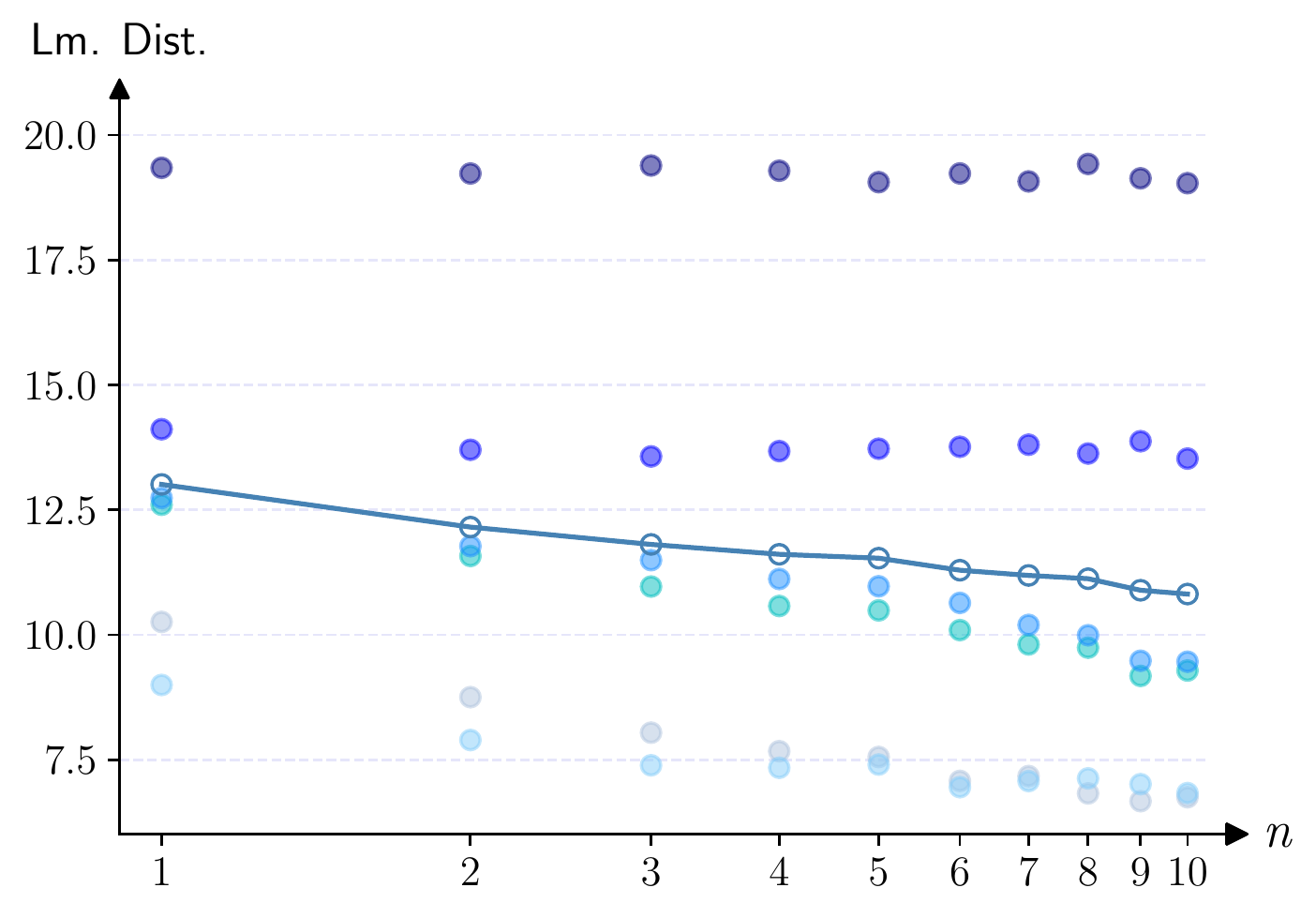}}
\subfigure[Dice scores on LPBA.]{ \label{brain_dice}
   \includegraphics[width=0.32\linewidth]{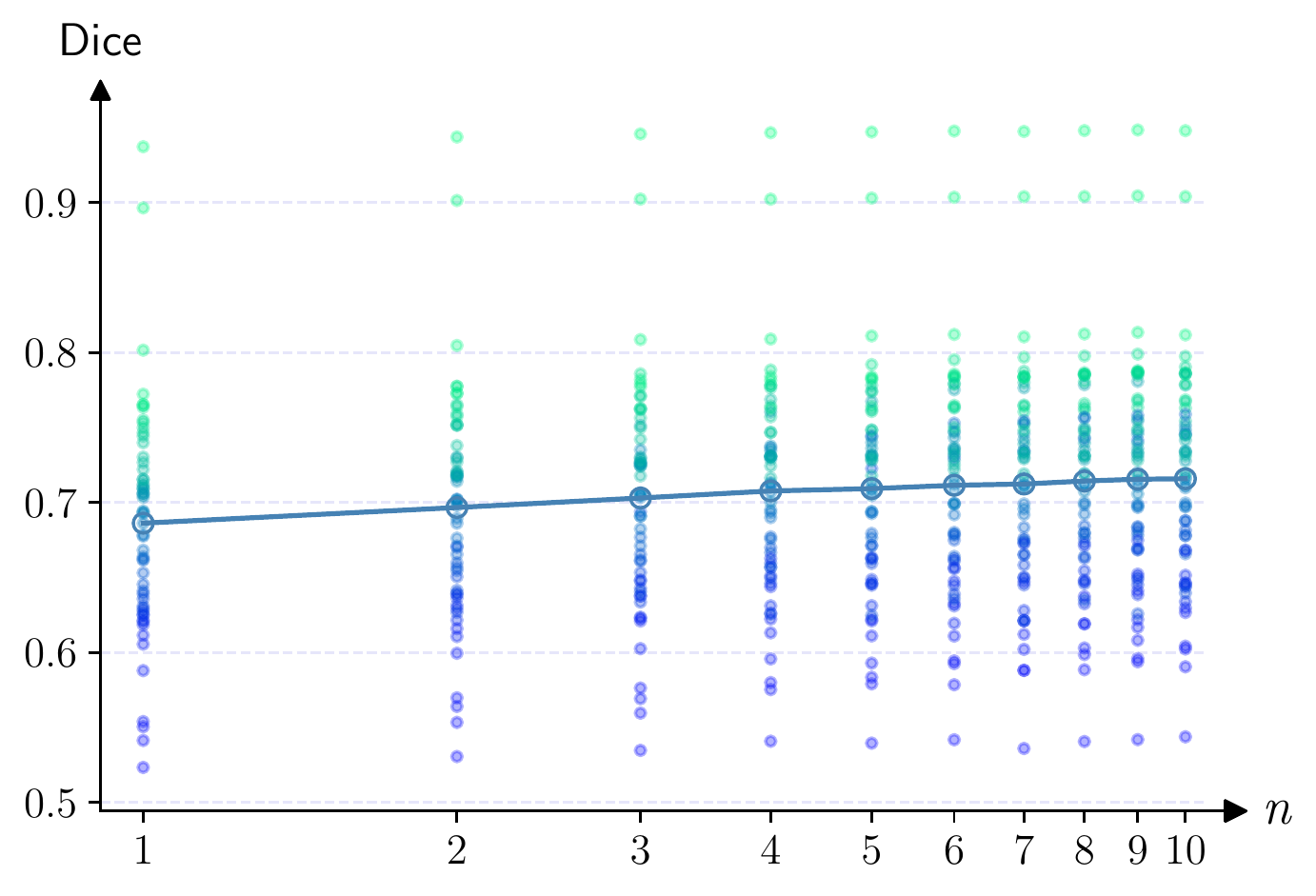}}
\end{center}
   \caption{Plot of our results with respect to the number of cascades ($n$) of the base network VTN, corresponding to the data in Table~\ref{table:cascade}. The $x$-axes are in log scale since it better reflects the trends. (a) plots the Dice scores evaluated on the liver datasets (SLIVER, LiTS, and LSPIG). (b) plots the landmark distances evaluated on SLIVER, while the distances (average across instances) of 6 landmarks are scattered with respective colors and the line stands for the mean values. (c) plots the Dice scores evaluated on the brain dataset LPBA, while the Dice scores (average across instances) of 56 anatomical structures are scattered respectively and the line stands for the mean values. }
\label{fig:plot}
\end{figure*}

\paragraph{Number of Cascades.}

Table \ref{table:cascade} presents the results with respect to different number of recursive cascades, choosing either VTN or VoxelMorph as our base network. As shown in the table, recursive cascaded networks achieve consistent performance gains independently of the base network. Our 3-cascade VTN (in Table \ref{table:cascade}) already outperforms VTN (ADDD) (in Table \ref{table:method}) although they have similar network architectures, mainly because our intermediate cascades learn progressive alignments better with only the similarity loss drawn on the final warped image. Figure \ref{fig:plot} plots our results for better illustrating the increasing trend.  Note that our architecture requires a linear time increment, but cascading a small-size base network like VTN is quite efficient.

\paragraph{Shared-Weight Cascading.}

Deeper cascades can be directly constructed using weight sharing. As we suggest, an $r$$\times$$n$-cascade network successively repeats each of the jointly trained $n$ cascades for $r$ times during testing. A linear time increment is also required. This technique ensures an increasing similarity between the warped moving image and the fixed image, but we note that it does not always get positive performance gains.

Table \ref{table:shared} presents the results of shared-weight cascaded networks, together with the image similarity (correlation coefficient is used in this paper). The image similarity is always increasing as we expect. Shallower cascaded networks benefit more from this technique relatively to the deeper ones, since the images are still not well-registered (with relatively low similarity, as shown in the table).
Less expected results on LiTS and LSPIG datasets may imply that this additional technique has a limited generalizability.

Note that shared-weight cascades generally perform worse than their jointly trained counterparts. More than 3 times of shared-weight cascades are very likely to deteriorate the quality (which partly coincides with previous studies), further proving the end-to-end learning to be vital.

\begin{table*}[t]
\scriptsize
\centering
\begin{threeparttable}
\begin{tabular}{|c||ccc|cc|cc||cc|}
\hline
Cascade & \multicolumn{3}{c|}{SLIVER} & \multicolumn{2}{c|}{LiTS} & \multicolumn{2}{c||}{LSPIG} & \multicolumn{2}{c|}{LPBA}  \\
& Dice  & Lm.\@ Dist. & Similarity & Dice & Similarity & Dice & Similarity & Avg.\@ Dice & Similarity     \\
\hline\hline
1$\times$1 & 0.914 (0.025) & 13.0 (4.8) & 0.7458 (0.0396) & 0.870 (0.048) & 0.7386 (0.0468) & 0.833 (0.054) & 0.7527 (0.0515) & 0.686 (0.014) & 0.9814 (0.0021) \\
2$\times$1 & 0.932 (0.020) & 12.6 (5.0) & 0.8108 (0.0289) & 0.886 (0.048) & 0.8045 (0.0376) & 0.840 (0.057) & 0.8162 (0.0392) & 0.694 (0.014) & 0.9845 (0.0016) \\
3$\times$1 & 0.937 (0.019) & 12.5 (5.1) & 0.8333 (0.0248) & 0.888 (0.050) & 0.8272 (0.0336) & 0.839 (0.057) & 0.8369 (0.0338) & 0.695 (0.013) & 0.9854 (0.0014) \\
4$\times$1 & 0.938 (0.018) & 12.5 (5.2) & 0.8444 (0.0227) & 0.887 (0.053) & 0.8381 (0.0314) & 0.837 (0.057) & 0.8467 (0.0305) & 0.692 (0.013) & 0.9857 (0.0011) \\
5$\times$1 & 0.939 (0.018) & 12.5 (5.2) & 0.8510 (0.0214) & 0.886 (0.056) & 0.8446 (0.0300) & 0.835 (0.058) & 0.8518 (0.0289) & 0.686 (0.013) & 0.9857 (0.0010) \\
\hline
1$\times$2 & 0.935 (0.020) & 12.2 (4.7) & 0.8270 (0.0297) & 0.891 (0.045) & 0.8209 (0.0367) & 0.843 (0.061) & 0.8435 (0.0369) & 0.697 (0.014) & 0.9854 (0.0017) \\
2$\times$2 & 0.947 (0.017) & 11.6 (4.8) & 0.8779 (0.0198) & 0.900 (0.049) & 0.8715 (0.0282) & 0.847 (0.063) & 0.8919 (0.0243) & 0.701 (0.014) & 0.9885 (0.0011)\\
3$\times$2 & 0.948 (0.016) & 11.5 (4.8) & 0.8930 (0.0171) & 0.900 (0.054) & 0.8865 (0.0254) & 0.845 (0.063) & 0.9039 (0.0211) & 0.697 (0.014) & 0.9895 (0.0008)\\
\hline
1$\times$3 & 0.943 (0.018) & 11.8 (4.7) & 0.8584 (0.0245) & 0.900 (0.045) & 0.8535 (0.0318) & 0.850 (0.060) & 0.8774 (0.0282) & 0.703 (0.014) & 0.9876 (0.0014) \\
2$\times$3 & 0.951 (0.015) & 11.2 (4.8) & 0.8977 (0.0168) & 0.905 (0.052) & 0.8927 (0.0246) & 0.852 (0.061) & 0.9102 (0.0210) & 0.710 (0.014) & 0.9904 (0.0009) \\
3$\times$3 & 0.951 (0.015) & 11.1 (4.9) & 0.9088 (0.0146) & 0.904 (0.058) & 0.9037 (0.0225) & 0.850 (0.062) & 0.9189 (0.0188) & 0.711 (0.014) & 0.9916 (0.0007)\\
\hline
1$\times$5 & 0.949 (0.015) & 11.5 (4.8) & 0.8926 (0.0186) & 0.908 (0.051) & 0.8893 (0.0254) & 0.853 (0.063) & 0.9088 (0.0223) & 0.709 (0.014) & 0.9894 (0.0010) \\
2$\times$5 & 0.954 (0.013) & 10.8 (4.9) & 0.9215 (0.0131) & 0.908 (0.061) & 0.9184 (0.0198) & 0.851 (0.063) & 0.9334 (0.0164) & 0.715 (0.013) & 0.9921 (0.0006) \\
3$\times$5 & 0.954 (0.013) & 10.6 (5.0) & 0.9308 (0.0115) & 0.906 (0.067) & 0.9278 (0.0182) & 0.845 (0.065) & 0.9406 (0.0145) & 0.715 (0.013) & 0.9930 (0.0005)\\
\hline
1$\times$10 & 0.953 (0.014) & 10.8 (4.9) & 0.9163 (0.0145) & \textbf{0.909 (0.060)} & 0.9129 (0.0211) & \textbf{0.855 (0.059)} & 0.9290 (0.0174) &0.716 (0.013) & 0.9918 (0.0008) \\
2$\times$10 & \textbf{0.956 (0.012)} & \textbf{10.2 (4.7)} & 0.9384 (0.0106) & 0.908 (0.070) & 0.9355 (0.0171) & 0.849 (0.062) & 0.9471 (0.0132) & \textbf{0.719 (0.012)} & 0.9942 (0.0005) \\
3$\times$10 & \textbf{0.956 (0.012)} & \textbf{10.2 (4.7)} & \textbf{0.9461 (0.0094)} & 0.905 (0.076) & \textbf{0.9434 (0.0158)} & 0.841 (0.068) & \textbf{0.9534 (0.0112)} & 0.717 (0.012) & \textbf{0.9951 (0.0004)} \\
\hline
\end{tabular}
\caption{Results of recursive cascaded networks built on the base network VTN, with different times (1$\times$, 2$\times$, 3$\times$, or more) of shared-weight cascades. 
Similarity is measured by the correlation coefficient between the warped moving image and the fixed image.}
\label{table:shared}
\end{threeparttable}
\end{table*}

\begin{table}[t]
\tiny
\centering
\begin{threeparttable}
\begin{tabular}{|c||cc|c|c||c|}
\hline
Architecture & \multicolumn{2}{c|}{SLIVER}  & LiTS          & LSPIG & LPBA \\
      & Dice          & Lm.\@ Dist.     & Dice          & Dice  & Avg.\@ Dice\\
\hline\hline
VoxelMorph & 0.913 (0.025) & 13.1 (4.7) & 0.867 (0.050) & 0.833 (0.057) & 0.688 (0.015) \\
VM x2 & 0.922 (0.024) & 13.0 (4.9) & 0.879 (0.047) & 0.839 (0.058) & 0.691 (0.015) \\
VM-double & 0.919 (0.025) & 12.9 (4.9) & 0.877 (0.048) & 0.833 (0.059) & 0.689 (0.015) \\
VM xx2 & 0.925 (0.023) & \textbf{12.8 (4.9)} & 0.881 (0.047) & 0.843 (0.057) & 0.693 (0.014) \\
2$\times$1-cascade VM & 0.930 (0.021) & \textbf{12.8 (4.8)} & 0.883 (0.051) & 0.840 (0.060) & 0.697 (0.014) \\
2-cascade VM & \textbf{0.933 (0.021)} & \textbf{12.8 (4.8)} & \textbf{0.888 (0.048)} & \textbf{0.845 (0.057)} & \textbf{0.699 (0.014)} \\
\hline
\end{tabular}
\caption{Comparison against other variants of VoxelMorph (VM), including VM x2 (doubling the feature counts of every convolutional layer), VM-double (doubling the number of convolutional layers at each level), and VM xx2 (doubling the encoder-decoder architecture cascade-like).}
\label{table:VMx2}
\end{threeparttable}
\end{table}

\begin{table}[t]
\tiny
\centering
\begin{threeparttable}
\begin{tabular}{|c||ccc|}
\hline
Method                          & \multicolumn{3}{c|}{Avg. Dice} \\
                                & Atlas1          & Atlas2     & Atlas3          \\
\hline\hline
ANTs SyN        & 0.708 (0.015) & 0.717 (0.011) & 0.707 (0.015) \\
Elastix B-spline & 0.675 (0.013) & 0.684 (0.011) & 0.670 (0.013) \\
\hline
VoxelMorph          & 0.688 (0.015) & 0.694 (0.010) & 0.678 (0.015)  \\
5-cascade VoxelMorph            & 0.708 (0.015) & 0.714 (0.011) & 0.702 (0.014) \\
3$\times$5-cascade VoxelMorph   & 0.715 (0.014) & 0.721 (0.012) & 0.713 (0.013) \\
\hline
VTN (ADDD) & 0.701 (0.014) & 0.709 (0.011) & 0.695 (0.015) \\
10-cascade VTN                  & 0.716 (0.013) & 0.723 (0.010) & 0.712 (0.013) \\
2$\times$10-cascade VTN         & \textbf{0.719 (0.012)} & \textbf{0.725 (0.011)} & \textbf{0.716 (0.013)} \\
\hline
\end{tabular}
\caption{Experiments on different atlases in LPBA.}
\label{table:atlas}


\end{threeparttable}
\end{table}

\paragraph{Cascades vs.\@ Channels vs.\@ Depth.}

VoxelMorph (VM)~\cite{ExVM} suggests that the number of channels in the convolutional layers can be doubled for a better performance. We compare this variant (VM x2) against the jointly trained 2-cascade VM as well as a shared-weight 2$\times$1-cascade VM, shown in Table \ref{table:VMx2}. VM x2 performs better than the original one as they suggest, but worse than both of our cascade methods. On the other hand, the number of parameters in VM x2 is 4 times as large as that in VoxelMorph (as well as 2$\times$1-cascade VM), and 2 times as large as that in 2-cascade VM.

However, one may wonder that whether simply deeper networks would do the trick. To this end, we construct VM-double by doubling the number of convolutional layers at each U-net level, and also an encoder-decoder-encoder-decoder architecture denoted VM xx2, which looks similar to a 2-cascade VM except the explicit warping. They have approximately the same amount of parameters compared to the 2-cascade VM, but are outperformed by a considerable margin.
This experiment implies that our improvements are essentially based on the proposed recursive cascade architecture rather than simply introducing more parameters.

\paragraph{Atlas Analysis.}

The performance for atlas-based registration may vary depending on the chosen atlas. As a comparison, we retrain the models on two more (the second and the third) atlases in the LPBA dataset, shown in Table \ref{table:atlas}. These results indicate that our performance is consistent and robust to the choice of atlas.

\section{Discussion}

Recursive cascaded networks are quite simple to implement, and also easy to train. We do not tune the ratios of losses when training more cascades, nor the training schedule, showing the robustness of our architecture. If more resources are available or a distributed learning platform is being used, we expect that the performance can be further improved by deeper cascades, and also, training or fine-tuning shared-weight cascades would be an alternative choice. A light-weight base network is also worth an exploration.

A possible limitation of this work would be on the smoothness of the composed field. Theoretically, recursive cascaded networks preserve the image topology as long as every subfield does. However, folding area is common in currently proposed methods and may be amplified during recursion, which brings challenges especially for the use of weight sharing techniques. This problem can be reduced by taking a careful look on the regularization terms, or designing a base network that guarantees invertibility.

\section{Conclusion}


We present a deep recursive cascade architecture and evaluate its performance in deformable medical image registration. Experiments based on diverse evaluation metrics demonstrate that this architecture achieves significant gains over state-of-the-art methods on both liver and brain datasets. With the superiority of good performance, the general applicability of the unsupervised method, and being independent of the base network, we expect that the proposed architecture can potentially be extended to all deformable image registration tasks.


{\small
\bibliographystyle{ieee_fullname}
\bibliography{ref}
}

\clearpage



\section*{Supplementary material}

# Recursive Cascaded Networks for Unsupervised Medical Image Registration
## Supplementary Material

## Abstract

*This supplementary material provides the visualization of some examples across all the evaluation datasets we use. Figure 1 presents an overview of the evaluation datasets. Figures 2, 3, 4, 5 show some examples on each of the liver datasets (SLIVER, LiTS, LSPIG) and the brain dataset (LPBA) respectively. We visualize these results by drawing the output flows, computing the warped images as well as the warped grids, and comparing the warped annotations with the fixed images.*

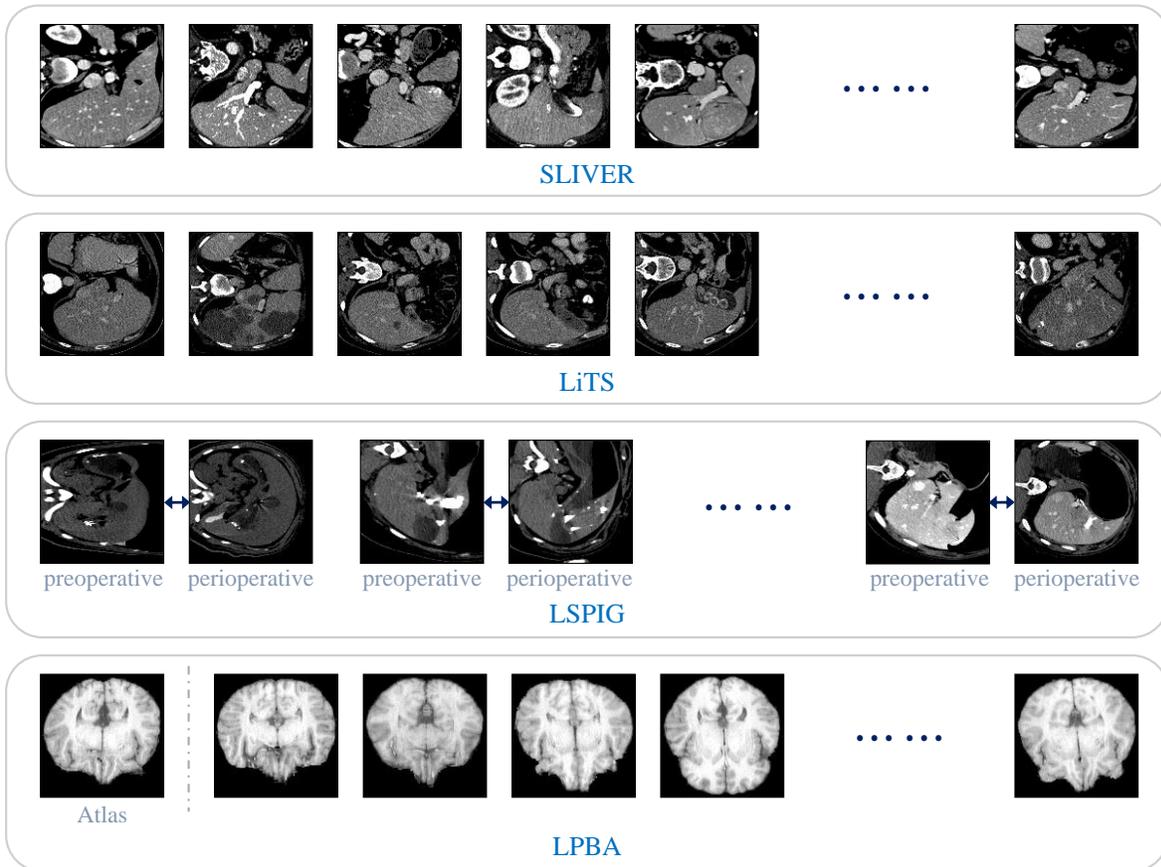

Figure 1. Samples of input images on each of the evaluation dataset. SLIVER, LiTS, LSPIG are for liver CT scans; LPBA is for brain MRIs. Median slices of those 3D images are shown. Every nonidentical image pair in SLIVER as well as LiTS is evaluated. Every image pair in LSPIG which comes from a same pig (preoperative to perioperative, or vice versa) is evaluated. Every image in LPBA registered to the fixed atlas is evaluated.

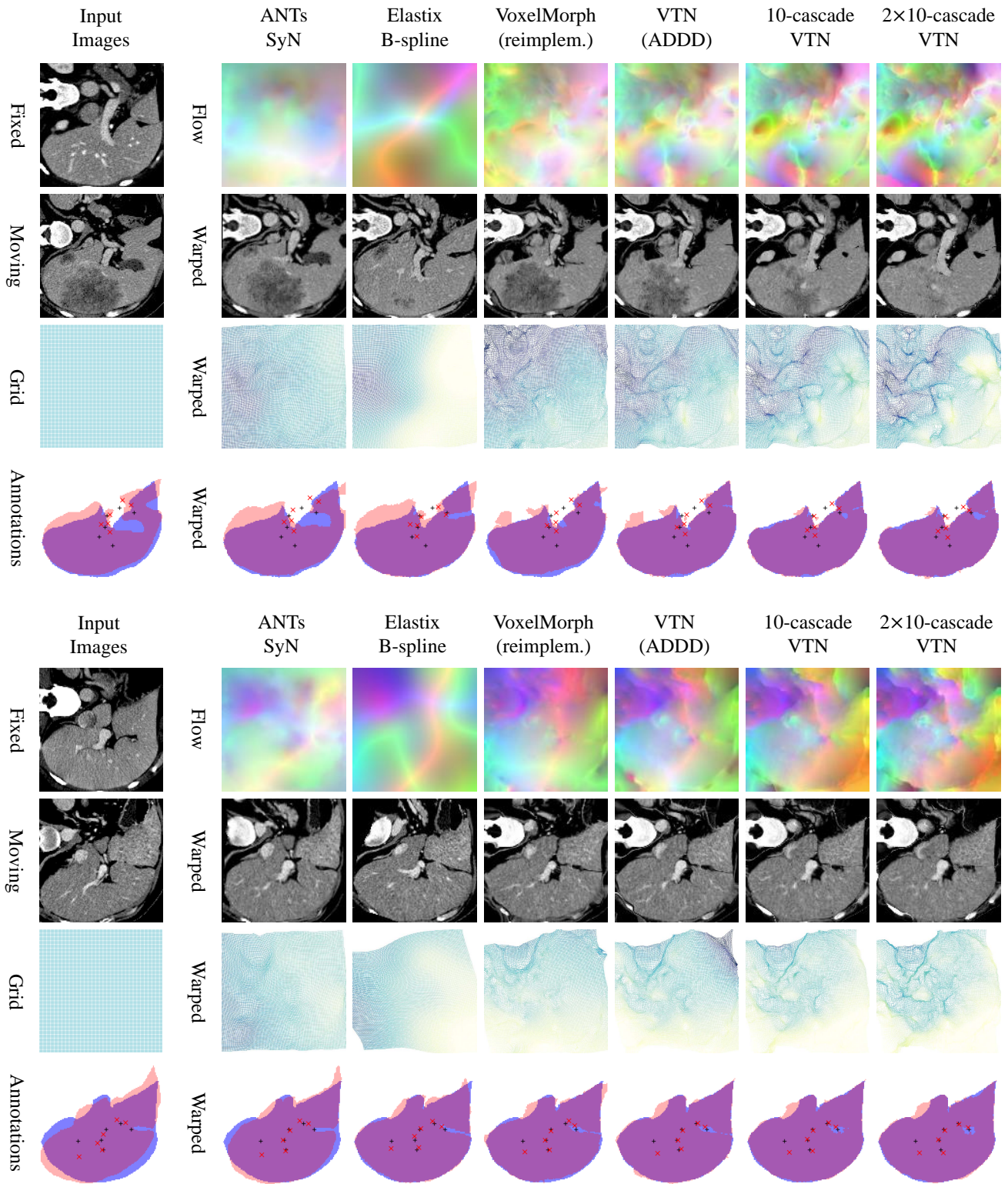

Figure 2. Examples across different methods on the liver dataset SLIVER. Flow fields are drawn by mapping the absolute value of the three components $(x, y, z)$ of flow displacements into color channels (R, G, B) respectively. Grids of deeper color represent lower height. Landmarks are projected onto the plane. Black plusses represent the landmarks of the fixed image, and red crosses represent the landmarks of the moving image or the warped images. Segmentations are sliced. Blue areas represent the segmentations of the fixed image, and red areas represent the segmentations of the moving image or the warped images. Coincident areas are in purple.

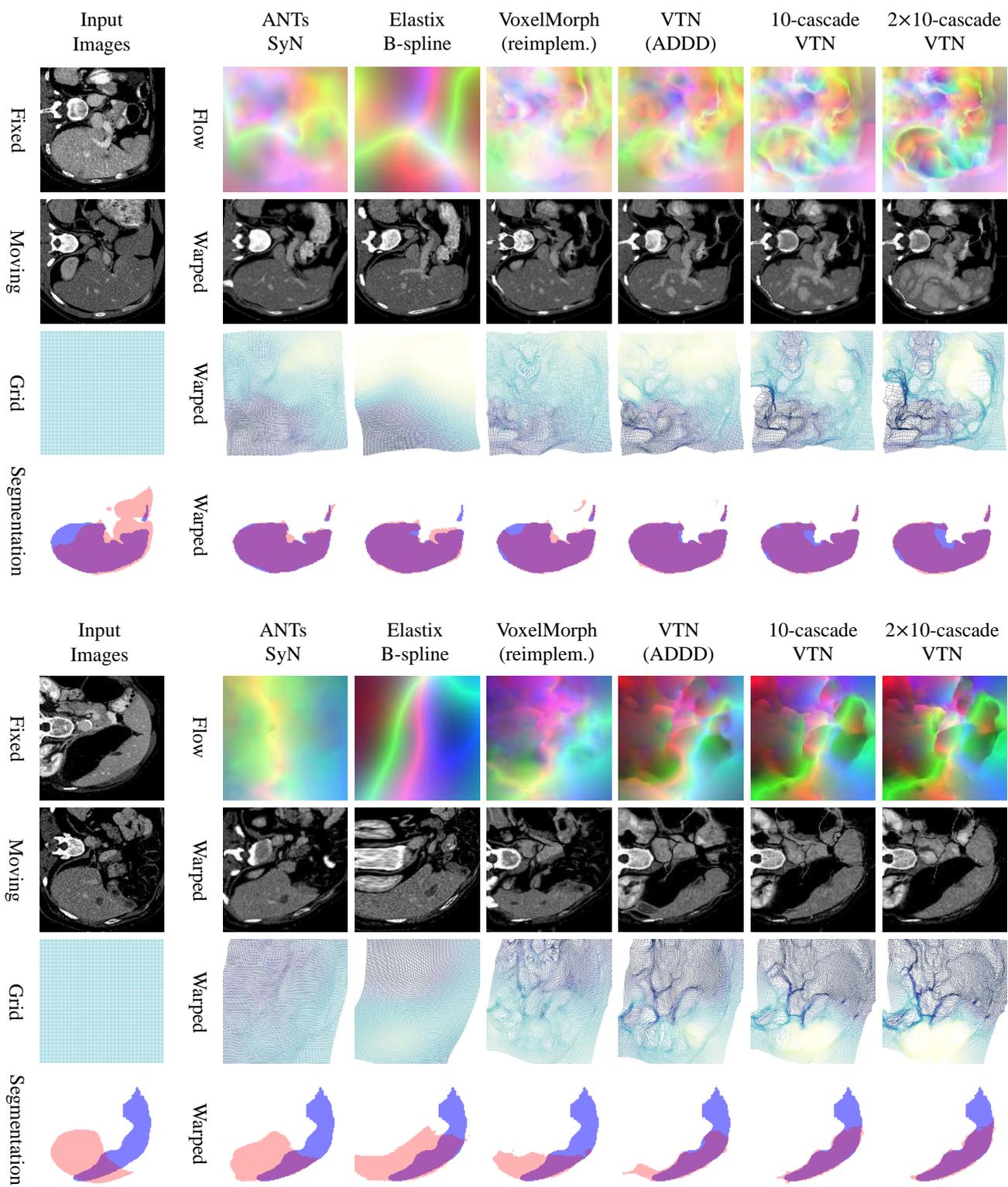

Figure 3. Examples across different methods on the liver dataset LiTS. Median slices of images, flows, and segmentations are shown.

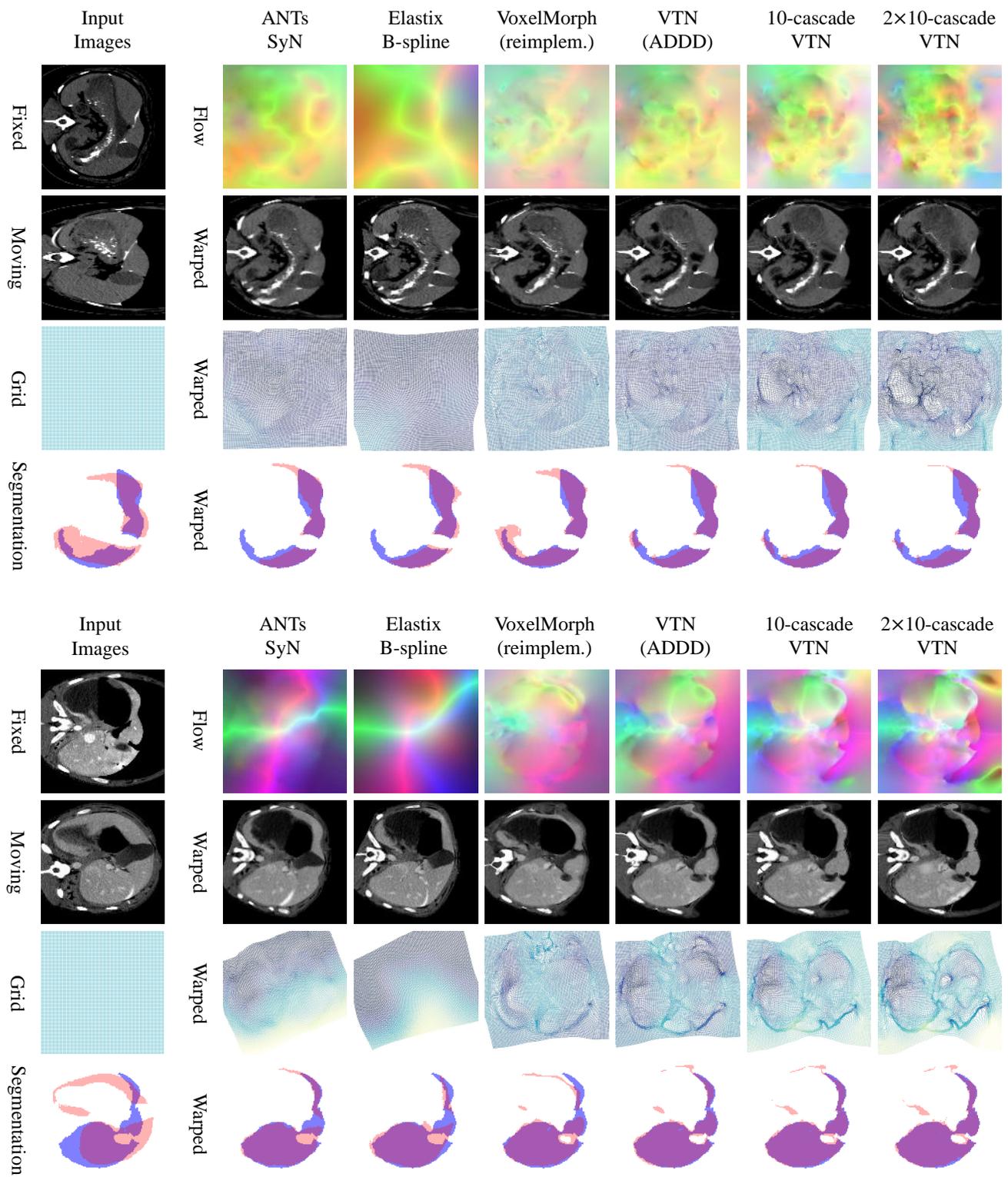

Figure 4. Examples across different methods on LSPIG, the liver dataset of pigs. Median slices of images, flows, and segmentations are shown.

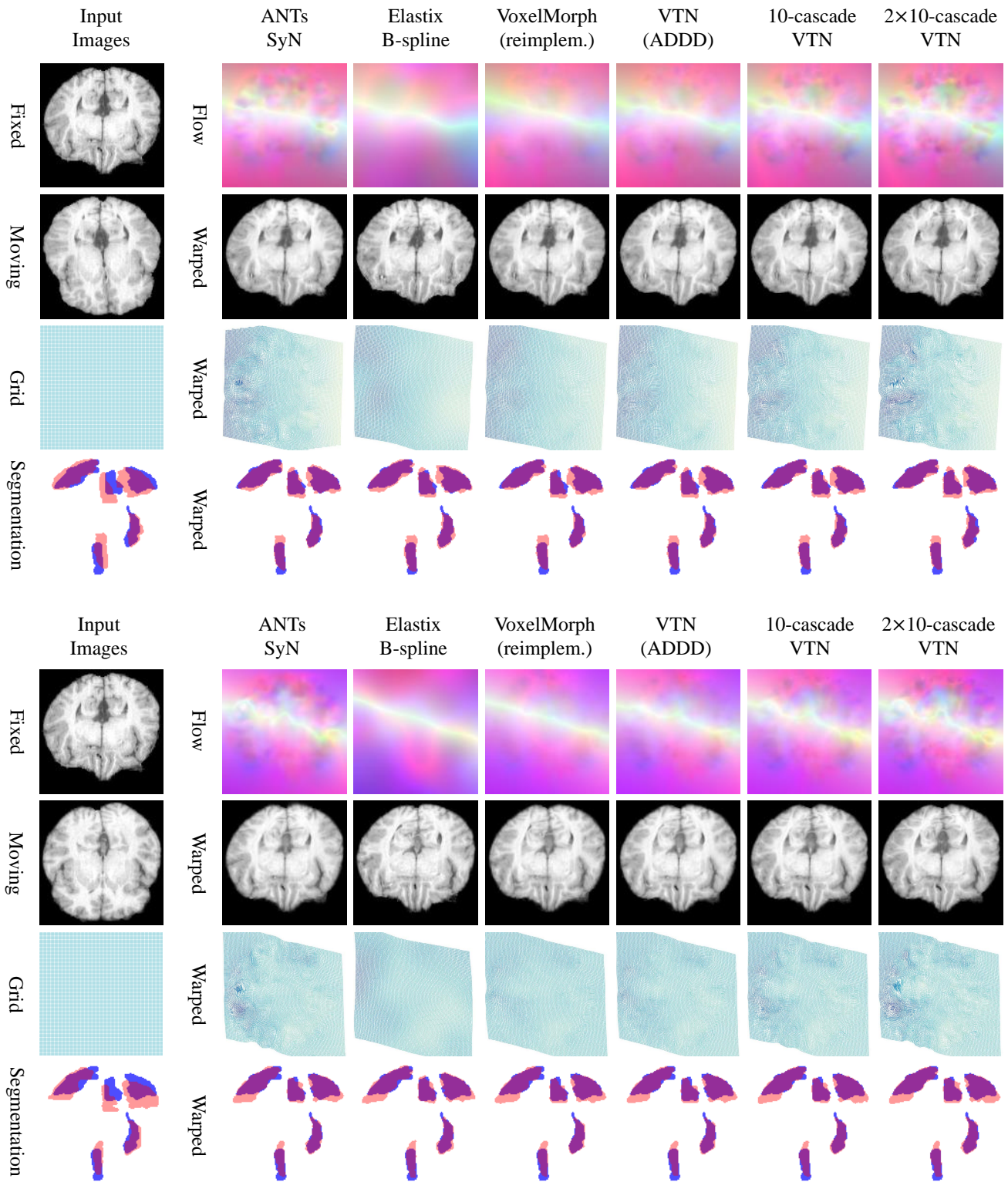

Figure 5. Examples across different methods on the brain dataset LPBA. Median slices of images and flows are shown. Segmentations of 5 chosen anatomical structures are projected onto the plane.



\end{document}